\documentclass[12pt,a4paper]{article}

\usepackage[top=2cm, bottom=3cm, left=2cm, right=2cm]{geometry}
\usepackage{color,colortbl}
\usepackage[table]{xcolor}
\usepackage{comment}

\usepackage[utf8]{inputenc}
\usepackage[T1]{fontenc}
\usepackage[english]{babel}

\usepackage{bm}
\usepackage{pifont}
\usepackage[normalem]{ulem}

\usepackage{amsthm,amsmath,amssymb}
\usepackage{esint}
\usepackage{mathrsfs}
\usepackage{esvect}
\usepackage{array}

\usepackage{numprint}
\usepackage{xlop} 

\usepackage{graphicx}
\usepackage{url}
\makeatletter
\g@addto@macro{\UrlBreaks}{\UrlOrds}
\makeatother

\usepackage{listings}
\usepackage{listingsutf8}
\usepackage{color}
\definecolor{mygreen}{RGB}{28,172,0}
\definecolor{mylilas}{RGB}{170,55,241}

\usepackage{algorithm}
\usepackage{algpseudocode}

\usepackage{fourier}
\usepackage{textcomp} 
\usepackage{stmaryrd}
\usepackage{ulem}
\usepackage{eurosym}
\usepackage{pifont}
\usepackage{fancybox,framed}
\usepackage{cancel}
\usepackage{lastpage}

\usepackage{pstricks} 
\usepackage{tikz}
\usepackage{chemist}

\usepackage{mathtools}
\usepackage{appendix}

\usepackage{hyperref}
\hypersetup{pdfstartview=XYZ}

\renewcommand{\leq}{\ensuremath{\leqslant}}
\renewcommand{\geq}{\ensuremath{\geqslant}}

\newcommand{\ceil}[1]{\left\lceil #1 \right\rceil}
\DeclareMathOperator*{\argmin}{arg\,min}

\usepackage{fancyhdr}
\usepackage{lastpage}
\usepackage{fontawesome}

\begin{document}

\definecolor{shadecolor}{gray}{0.9}
\begin{center}
\begin{shaded}
{\Large \textbf{ASTRIDE: Adaptive Symbolization for Time Series Databases}}
\end{shaded}
\end{center}

\bigbreak

\noindent\textbf{Sylvain W.\ Combettes}\textsuperscript{ 1}\hfill\href{mailto:sylvain.combettes@ens-paris-saclay.fr}{\ttfamily sylvain.combettes@ens-paris-saclay.fr}

\medbreak
\textbf{Charles Truong}\textsuperscript{1}\hfill\href{mailto:charles.truong@ens-paris-saclay.fr}{\ttfamily charles.truong@ens-paris-saclay.fr}

\medbreak
\textbf{Laurent Oudre}\textsuperscript{1}\hfill\href{mailto:laurent.oudre@ens-paris-saclay.fr}{\ttfamily laurent.oudre@ens-paris-saclay.fr}

\medbreak
{\small\textsuperscript{1}\emph{Université Paris-Saclay, Université Paris Cité, ENS Paris-Saclay, CNRS, SSA, INSERM, Centre Borelli\\
4, avenue des Sciences, Gif-sur-Yvette, F-91190, France}}

\bigbreak
\begin{center}
\begin{minipage}{0.9\linewidth}
\begin{center}
    \textbf{Abstract}
\end{center}
\vspace{-3pt}

We introduce ASTRIDE (Adaptive Symbolization for Time seRIes DatabasEs), a novel symbolic representation of time series, along with its accelerated variant FASTRIDE (Fast ASTRIDE). 
Unlike most symbolization procedures, ASTRIDE is adaptive during both the segmentation step by performing change-point detection and the quantization step by using quantiles.
Instead of proceeding signal by signal, ASTRIDE builds a dictionary of symbols that is common to all signals in a data set.
We also introduce D-GED (Dynamic General Edit Distance), a novel similarity measure on symbolic representations based on the general edit distance.
We demonstrate the performance of the ASTRIDE and FASTRIDE representations compared to SAX (Symbolic Aggregate approXimation), 1d-SAX, SFA (Symbolic Fourier Approximation), and ABBA (Adaptive Brownian Bridge-based Aggregation) on  reconstruction and, when applicable, on classification tasks.
These algorithms are evaluated on 86 univariate equal-size data sets from the UCR Time Series Classification Archive.
An open source GitHub repository called \texttt{astride} is made available to reproduce all the experiments in Python.

\bigbreak
\end{minipage}
\end{center}

\bigbreak

\section{Introduction}\label{sec:intro}

Over the past decades, the increasing amount of available time series data has led to a rising interest in time series data mining.
In many applications, the collected data take the form of complex time series which can be multivariate, multimodal, or noisy.
A fundamental issue is to adopt an actionable representation which takes into account temporal information. In this regard, symbolic representations constitute a tool of choice~\cite{2007linexpsax}.
Symbolic representations of time series are used for data mining tasks such as classification~\cite{2007linexpsax, 2013_senin_saxvsm, 2014_schafer_boss, Nguyen2019InterpretableTS}, clustering~\cite{2007linexpsax}, indexing~\cite{2007linexpsax, Camerra2010iSAX2I}, anomaly detection~\cite{2020elsworthabba, 2022_chen_fabba}, motif discovery~\cite{2018_senin_grammarviz}, and forecasting~\cite{Elsworth2020TimeSF}.
The domain applications include finance~\cite{2006lkhagvaextended, 2012barnaghiensax}, healthcare~\cite{2010santannaclust}, and manufacturing~\cite{Park2020SAXARMDE}.

Briefly, most symbolization techniques follow two steps: a segmentation step where a real-valued signal $\bm{y} = (y_1, \ldots, y_n)$ of length $n$ is split into $w$ segments, then a quantization step where each segment is mapped to a discrete value $\hat{y}_i$ taken from a set $\{\text{\texttt{a}}_1, \ldots, \text{\texttt{a}}_A\}$ of $A$ symbols.
The resulting symbolic representation is the discrete-valued signal (or \emph{symbolic sequence}) $\hat{\bm{y}} = (\hat{y}_1, \ldots, \hat{y}_w)$.
The set of symbols $\{\text{\texttt{a}}_1, \ldots, \text{\texttt{a}}_A\}$ is usually called an \emph{alphabet} or \emph{dictionary}, and $A$ is the \emph{alphabet size}; the length $w$ of the symbolic representation is called the \emph{word length}.
While there exist many high-level representations for time series~\cite{2011reviewfu}, the two main advantages of symbolic representations are reduced memory usage, and often a better score on data mining tasks thanks to the smoothing effect induced by compression~\cite{2007linexpsax}.

In the present paper, we introduce \emph{ASTRIDE (Adaptive Symbolization for Time seRIes DatabasEs)}, a novel symbolic representation of time series data bases, along with its accelerated variant \emph{FASTRIDE (Fast ASTRIDE)}.
Unlike most symbolization techniques, ASTRIDE is adaptive during both the segmentation step by performing change-point detection and the quantization step by using quantiles.
As the segmentation and quantization are performed on the whole data set, a notable benefit of ASTRIDE is to define a common dictionary of symbols for all signals in the data set under consideration, thus further reducing memory usage.
ASTRIDE comes with \emph{D-GED (Dynamic General Edit Distance)}, a new similarity measure for symbolic sequences which is based on the general edit distance. As we shall see, ASTRIDE provides an intuitive symbolic representation which outperforms the state of the art in  classification accuracy and achieves competitive results in signal reconstruction.

The remainder of the paper is organized as follows.
Section~\ref{sec:background} provides an overview of symbolic representations and their similarity measures, highlights their limits, and presents our main contributions.
Section~\ref{sec:descriptionastride} introduces the novel ASTRIDE and FASTRIDE symbolic representations, as well as the new D-GED similarity measure.
Section~\ref{sec:expresults} contains an experimental evaluation of the accuracy of ASTRIDE and FASTRIDE for classification and for signal reconstruction compared to several state-of-the-art symbolization methods.
Section~\ref{sec:conclusion} provides concluding remarks.

\section{Background and motivations}\label{sec:background}

This section gives an overview of symbolic representations and their similarity measures, then assesses their limits and presents our contributions. It also provides a summary of some symbolization methods in Table~\ref{tab:symb_summary}.

\subsection{Overview of symbolic representations}\label{subsec:background}

\begin{figure}[hbtp]
    \centering
    \includegraphics[width=\linewidth]{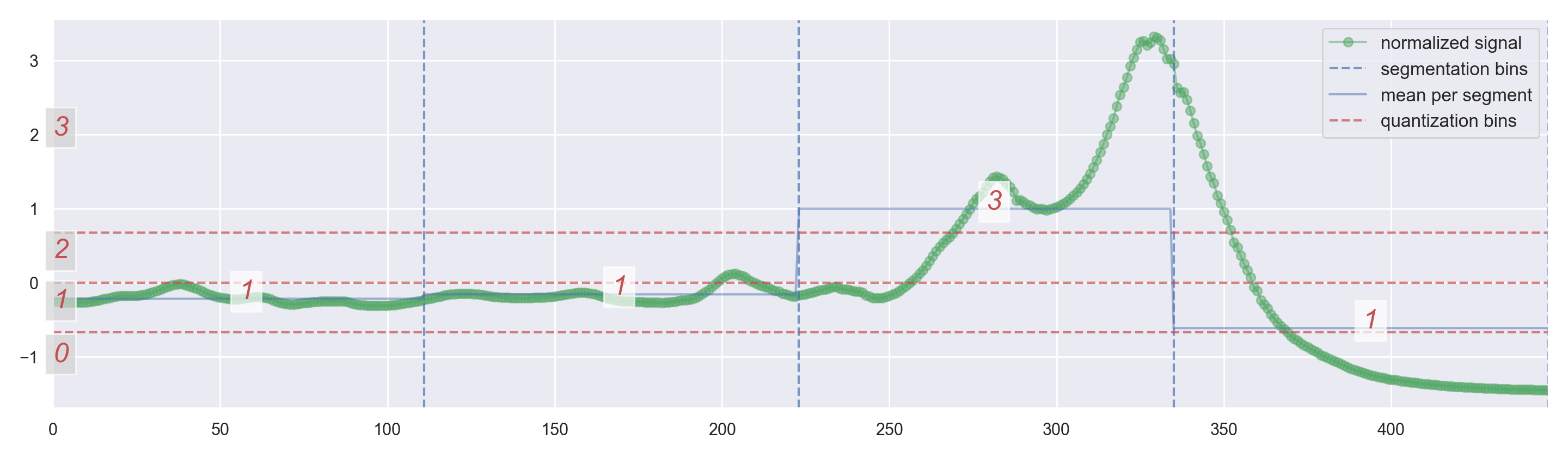}
    \includegraphics[width=\linewidth]{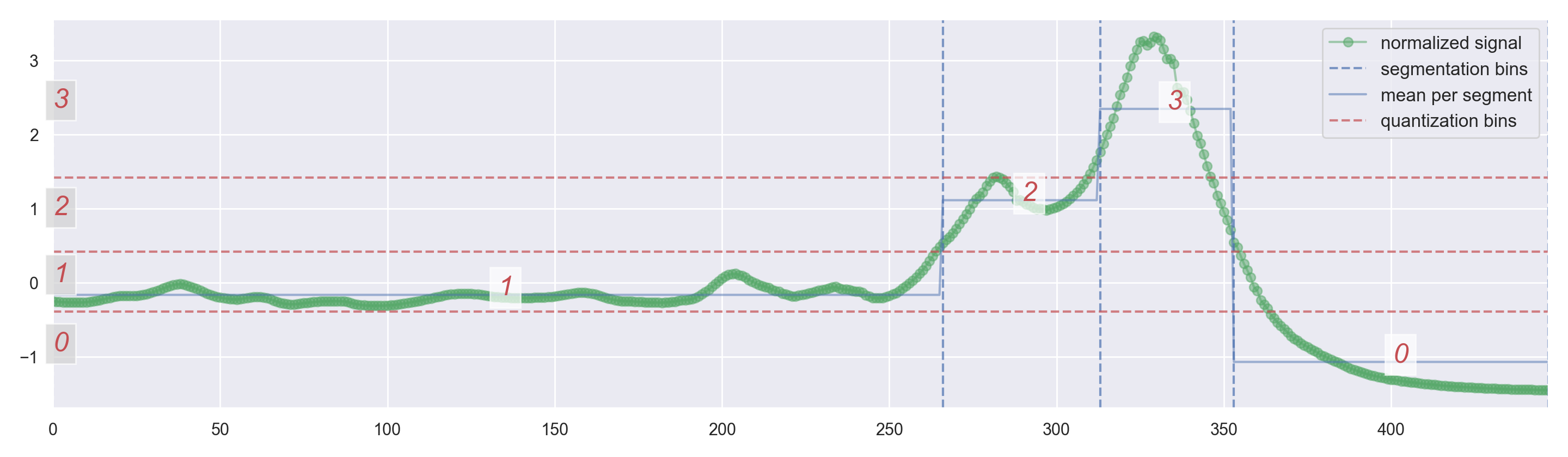}
    
    \caption{Example of a SAX (top) and ASTRIDE (bottom) representations of a signal from the Meat data set (UCR Time Series Classification Archive).
    The original length of the signal is $n=448$, and we use $w=4$ and $A=4$.
    The resulting symbolic sequence is \texttt{1131} for SAX, and \texttt{1230} for ASTRIDE. (ASTRIDE is described in Section~\ref{sec:descriptionastride}.)
    }
    \label{fig:symbolization_example}
\end{figure}

In 2003, a popular symbolic representation for time series was introduced: \emph{Symbolic Aggregate approXimation (SAX)}~\cite{2003linsax, 2007linexpsax}.
In SAX, a symbolic sequence is a chain of characters, for example \texttt{abbcaabc} (or \texttt{01120012}). SAX has two parameters: the word length $w$ and the alphabet size $A$.
For instance, in the symbolic sequence \texttt{abbcaabc}, the parameters are $w=8$ (length of the sequence) and $A=3$ (number of possible symbols).
The larger $w$ and $A$, the better the quality of the SAX representation, but the lower the compression.
Optimal values of $w$ and $A$ are highly dependent on the application and the data set.
In SAX, each signal is centered and scaled to unit variance, then split into $w$ segments of equal length.
Next, the means of all segments are grouped together in bins and each segment is represented by the bin where its mean falls into.
The bin boundaries are chosen so that all symbols are equiprobable under the assumption that the means follow a standard Gaussian distribution.
A SAX transformation of a signal taken from the UCR Time Series Classification Archive~\cite{UCRArchive2018} is shown in Figure~\ref{fig:symbolization_example}.

Since the introduction of SAX, many variants and symbolization techniques have been proposed.
First of all, some variants focus on the feature(s) per segment.
\emph{Extended SAX (ESAX)}~\cite{2006lkhagvaextended} represents each segment by its mean, minimum, and maximum values.
\emph{ENhanced SAX (EN-SAX)}~\cite{2012barnaghiensax} builds a vector for each segment with the minimum, maximum, and mean values.
These vectors are then clustered and a symbol is attributed to each cluster. 
\emph{Trend-based and Valued-based Approximation (TVA)}~\cite{2012esmaelmultivariate} discretizes the means and the trends (\texttt{U} for upwards, \texttt{D} for downwards, or \texttt{S} for straight trend).
Note that TVA can be applied to multivariate signals.
\emph{1d-SAX}~\cite{2013malinowski1dsax} represents two features with only one symbol per segment. It uses linear regression to compute the mean and the slope of each segment, then discretizes the mean (in $A_\text{mean}$ symbols) and the slope (in $A_\text{slope}$ symbols) separately using the same Gaussian assumption as in SAX. The final segment symbol is the combination of the mean symbol and the slope symbol.
The alphabet size is therefore $A = A_\text{mean} \cdot A_\text{slope}$.

Some symbolization procedures perform an non-uniform segmentation in order to better adjust to the signal.
Adaptive Segmentation Based Symbolic Representations (SBSR)~\cite{2006hugueneyadapt} can be viewed as a symbolic version of the \emph{Adaptive Piecewise Constant Approximation (APCA)} representation~\cite{2001chakrabartiapca}, just as SAX can be viewed as the symbolic version of the \emph{Piecewise Aggregate Approximation (PAA)}~\cite{2000keoghpaa, 2000yipaa}.
In SBSR, segment lengths adapt to the shape of the signal.
The \emph{Persist algorithm}~\cite{2005morchenadapt} is based on the persistence score of segments in a signal to detect changes in the signal, given a number of segments $w$. The more likely it is to observe the same segment as the previous one, the larger the persistence score.

Some symbolic representations have an adaptive quantization step in order to relax the Gaussian assumption on the data.
\emph{Genetic Algorithms-based SAX (GASAX)}~\cite{2012fuadgasax} works as SAX, but the bin boundaries are determined through a genetic algorithm (which is a class of optimization procedures). GASAX does not require any specific distribution of the data and there is no need to normalize it.
\emph{Adaptive SAX (aSAX)}~\cite{2010phamhotsax} uses a uniform segmentation and $K$-means clustering for the quantization.
A symbol-based procedure to detect phases of gait signals~\cite{2010santannaclust} uses piecewise linear segmentation then $K$-means clustering on the features per segment to get the symbols. The features per segment include slope, length, mean, and variance. Symbolization is used in order to have a interpretable representation of gait signals. However, no similarity measure is derived from this representation.
\emph{Adaptive Brownian Bridge-based Aggregation (ABBA)}~\cite{2020elsworthabba} is adaptive for both the segmentation and quantization steps.
It also chooses $w$ and $A$ through a data-driven procedure.
For the segmentation, adaptive piecewise linear continuous approximation of the signal is used. Each linear piece is chosen given a user-specified tolerance $tol$: when the value of $tol$ increases, the resulting number of segments $w$ decreases.
The quantization step consists in a $K$-means clustering of the tuples of the increment over the segment and the segment length, where the number of clusters is set to $A$.
ABBA uses a scaling parameter $scl$ that calibrates the importance of the length in relation to its increment: the clustering is performed on the increments alone for $scl = 0$, while the clustering is done on both the length and increment with the same importance when $scl = 1$.
Hence, the input parameters are the tolerance $tol$, the scaling $scl$, and the alphabet size $A$.
Note that when $A$ is not set by the user, ABBA does several runs the $K$-means algorithm to get the optimal value of $A$, resulting in a higher computational cost.
ABBA focuses on signal reconstruction: there is no mention of a similarity measure on symbolic representations.
\emph{Reconstruction} is the inverse transformation: the original signal is inferred from its transformation which is its symbolic sequence.
A recent faster variant of the ABBA method, \emph{fABBA}~\cite{2022_chen_fabba}, replaces the $K$-means clustering by a sorting-based aggregation procedure that does not require the user to specify $A$.

While previously mentioned methods symbolize each signal independently, some procedures operate on a data set of signals.
These methods share a dictionary of symbols across all signals of the considered data set.
\emph{Symbolic Fourier Approximation (SFA)}~\cite{2012_schafer_sfa} is based on the Discrete Fourier transform (DFT).
First, SFA selects the $w$ Fourier coefficients of lowest frequencies, and second, uses a procedure called \emph{Multiple Coefficient Binning (MCB)} to quantize them.
In detail, MCB computes a user-defined number $A$ of quantiles per Fourier coefficient across all signals of a data set, and each Fourier coefficient is represented by the bin (based on quantiles) to which it belongs.
In a supervised data mining task, the MCB bins are learned on a training set.
SFA naturally provides a low-pass filtering that reduces the influence of noise.
Also, no distance on SFA's symbolic representations is described.
Note that SFA does not go through a segmentation step, but still has the $w$ parameter that determines the length of the symbolic sequences.

Table~\ref{tab:symb_summary} summarizes the main SAX variants as well as our novel ASTRIDE and FASTRIDE representations that will be presented in Section~\ref{sec:descriptionastride}.

\subsection{Overview of similarity measures on symbolic sequences}\label{subsec:distsymb}

In order to use the learned symbolic representations for tasks such as classification or clustering, it is crucial to define a similarity measure between symbolic sequences, which can be viewed as character strings.
Defining an informative measure is a challenge that has received a lot of attention.

SAX employs MINDIST, a similarity measure on symbolic sequences.
Let $\bm{s} = (s_1, \ldots, s_n)$ and $\bm{t} = (t_1, \ldots, t_n)$ be two real-valued time series with $n$ samples. The Euclidean distance between $\bm{s}$ and $\bm{t}$ is given by
\begin{equation}
  \mathrm{EUCLIDEAN}(\bm{s}, \bm{t}) = \|\bm{s}-\bm{t}\|_2 = \sqrt{\sum_{i=1}^n \lvert s_i-t_i\rvert ^2}.
\end{equation}
The \emph{MINDIST} similarity measure between the resulting symbolic sequences $\hat{\bm{s}}$ and $\hat{\bm{t}}$ mimics the Euclidean distance
\begin{equation}
  \mathrm{MINDIST}\left(\hat{\bm{s}}, \hat{\bm{t}}\right) = \sqrt{\frac{n}{w}} \sqrt{\sum_{i=1}^w \lvert \mathrm{dist}\left(\hat{s}_i, \hat{t}_i\right)\rvert^2}
\end{equation}
where the function $\mathrm{dist()}$, based on a so-called look-up table, is illustrated in Table~\ref{tab:mindist}.
MINDIST requires the symbolic sequences to be of equal length.
\begin{table}[hbtp]
    \centering
    \begin{minipage}{0.8\linewidth}
    \caption{Example of look-up table for MINDIST with $A=4$. For example, $\mathrm{dist}(\text{\texttt{a}}, \text{\texttt{d}}) = 1.34$.}
    \label{tab:mindist}

    \centering
    \medbreak
    \begin{tabular}{@{}ccccc@{}}
        \hline
         & \texttt{a} & \texttt{b} & \texttt{c} & \texttt{d} \\
         \hline
        \texttt{a} & $0$ & $0$ & $0.67$ & $1.34$ \\
        \texttt{b} & $0$ & $0$ & $0$ & $0.67$ \\
        \texttt{c} & $0.67$ & $0$ & $0$ & $0$ \\
        \texttt{d} & $1.34$ & $0.67$ & $0$ & $0$ \\
        \hline
    \end{tabular}
    \end{minipage}
\end{table}
For a given value of the alphabet size $A$, this table is calculated only once, and then stored for fast look-up.
For all look-up tables, whatever the alphabet size, the value in the cell of indexes $(i,j)$ is given by
\begin{equation}
    \mathrm{cell}_{i,j} = \begin{cases}
        0, \quad \text{if}\ \lvert i-j\rvert \leq 1 \\
        \beta_{\max(i,j)-1}-\beta_{\min(i,j)}, \quad \text{otherwise}
    \end{cases}
\end{equation}
where the $\beta_k$ are the boundaries of the bins used by SAX to discretize the segment means.
MINDIST is not a true metric, as $\mathrm{dist}(\text{\texttt{a}}, \text{\texttt{b}})=0$ for example (see Table~\ref{tab:mindist}).
The MINDIST similarity measure on SAX symbolic sequences lower bounds the Euclidean distance on original signals, i.e.,
\begin{equation}
    \label{eq:lb}
   \mathrm{MINDIST}\left(\hat{\bm{s}}, \hat{\bm{t}}\right) \leq \mathrm{EUCLIDEAN}(\bm{s}, \bm{t}).
\end{equation}
This property implies that the similarity matching in the reduced space maintains its meaning with regards to the original space.
In turn, property~\eqref{eq:lb} ensures exact indexing of data in the sense that there will be no false negatives.

Most symbolization methods that possess a similarity measure are variants of MINDIST, but often lose the lower bounding property~\eqref{eq:lb}.
Note that EN-SAX~\cite{2012barnaghiensax} uses cosine similarity between vectors of features per segment as the similarity measure, so this similarity measure is not defined directly on the symbolic sequences but on their intermediate representation.
SBSR~\cite{2006hugueneyadapt} explores similarity measures that have additional information than MINDIST, namely the data set global information, the time series local information, and the episode local information.

The literature on symbolic representations advocates that a major advantage of symbolic representations is their ability to leverage the richness of the bioinformatics and text processing communities~\cite{2007linexpsax, 2020elsworthabba}.
However, to the best of our knowledge, existing symbolic representations do not seem to use similarity measures defined on strings.
A review on string matching is given in~\cite{2001navarrostring}.
A popular similarity measure is the edit distance: for two strings \texttt{s1} and \texttt{s2}, it is the minimal cost of a sequence of operations that transform \texttt{s1} into \texttt{s2}. The edit distance is also called the Levenshtein distance~\cite{levenshtein1966binary}.
The allowed operations are the following:
\begin{itemize}
    \item Insertion of a simple character in a string.
    \item Deletion of a simple character in a string.
    \item Substitution of simple characters (with a different one) in both strings.
\end{itemize}
To each operation corresponds a cost, and this cost also depends on the characters involved.
The edit distance can compare symbolic sequences of different lengths, because it allows for insertions and deletions.
The cost of a sequence of operations is the sum of the costs of the simple operations.
If all the simple operations have a cost of $1$, whatever the operation or the characters involved, it is called the \emph{simple edit distance}.
The simple edit distance is the minimum number of insertions, deletions and substitutions to make both strings equal.
If the three authorized operations have different costs or the costs depend on the characters involved, it is called the \emph{general edit distance}.
The simple edit distance $\mathrm{ed()}$ satisfies the following properties: $\mathrm{ed}(s_1, s_1)=0$, $\mathrm{ed}(s_1, s_2)>0$ if $s_1\neq s_2$ (positivity), $\mathrm{ed}(s_1, s_2) = \mathrm{ed}(s_2, s_1)$ (symmetry), and $\mathrm{ed}(s_1, s_2) \leq \mathrm{ed}(s_1, s_3) + \mathrm{ed}(s_3, s_2)$ (triangle inequality) for all strings $s_1$, $s_2$, and $s_3$.
Hence, the simple edit distance is a metric.
Moreover, we have $0 \leq \mathrm{ed}(s_1, s_2) \leq \max(\lvert s_1\rvert, \lvert s_2 \rvert)$ for all strings $s_1$ and $s_2$.

In bioinformatics, many algorithms are used to align DNA sequences of nucleotides, meaning strings composed of the letters \texttt{A}, \texttt{C}, \texttt{G}, and \texttt{T}.
These algorithms often use the edit distance as a score.
Exhaustive searches look for all possible alignments and retrieve the alignment(s) with the optimal score, which is very costly.
An alternative is the Needleman-Wunsch algorithm~\cite{needleman1970general} which uses dynamic programming for global sequence alignment. It determines the optimal alignment of all possible prefixes of the first sequence with all possible prefixes of the second sequence by going from the smallest to the largest prefixes.
The Smith and Waterman algorithm~\cite{smith1981identification} is a variation of the Needleman-Wunsch algorithm which performs local sequence alignment.

For each symbolization method presented in Section~\ref{subsec:background}, Table~\ref{tab:symb_summary} indicates whether it comes with a compatible similarity measure.

\begin{table}[hbtp]\small
\centering
\caption{
{\small Synthetic comparison of some symbolization methods (including our proposed ASTRIDE and FASTRIDE) for a word length $w$ and an alphabet size $A$, with a data set composed of $N$ signals whose values are encoded on $n_\text{bits}$ bits (for example $n_\text{bits}=32$ bits or $n_\text{bits}=64$ bits.) The $A_\text{mean}$, $A_\text{min}$, $A_\text{max}$, $A_\text{slope}$, $A_\text{clusters}$, $A_\text{coefficients}$ are respectively the number of symbols used to encode the mean, the minimum, the maximum, the slope, the number of clusters, and the number of Fourier coefficients.\\
$^{\dagger}$For ESAX, $A=A_{\text{mean}}=A_{\text{min}}=A_{\text{max}}$.\\
$^{\ddagger}$For SFA, there is no segmentation.\\
$^{\S}$For ASTRIDE, the lengths are common to all signals.}
}

\smallbreak
\label{tab:symb_summary}
\rotatebox{90}{\begin{tabular}[\textwidth]
{@{}p{0.13\textwidth}p{0.15\textwidth}p{0.08\textwidth}p{0.08\textwidth}p{0.08\textwidth}p{0.22\textwidth}p{0.17\textwidth}p{0.08\textwidth}@{}}
    \hline 
    SAX variant & Features used per segment & Adaptive segmentation? & Adaptive quantization? & Shared dictionary? & Space complexity (in bits) for a single symbolic sequence & Space complexity (in bits) of the dictionary of symbols for a data set of $N$ signals & Similarity measure? \\
    \hline
    SAX~\cite{2003linsax} & means & \ding{56} & \ding{56} & \ding{52} & $w \ceil{\log_2(A_{\text{mean}})}$ & $n_\text{bits} A_{\text{mean}}$ & \ding{52} \\[1ex]
    ESAX~\cite{2006lkhagvaextended} & maxs, mins, means & \ding{56} & \ding{56} & \ding{52} & $w \ceil{\log_2(A_{\text{mean}}A_{\text{min}}A_{\text{max}})}$ & $n_\text{bits} A^{\dagger}$ & \ding{52} \\[1ex]
    EN-SAX~\cite{2012barnaghiensax} & maxs, mins, means (as vectors) & \ding{56} & \ding{52} & \ding{56} & $w \ceil{\log_2(A_{\text{clusters}})}$ & $3Nn_\text{bits}A_{\text{clusters}}$ & \ding{52} \\[1ex]
    TVA~\cite{2012esmaelmultivariate} & means, directions & \ding{56} & \ding{56} & \ding{52} & $w \ceil{\log_2(A_\text{mean}  A_\text{slope})}$ & $n_\text{bits} (A_\text{mean} + A_\text{slope})$ & \ding{56} \\[1ex]
    1d-SAX~\cite{2013malinowski1dsax} & means, slopes & \ding{56} & \ding{56} & \ding{52} & $w \ceil{\log_2(A_{\text{mean}}A_{\text{slope}})}$ & $n_\text{bits} A_{\text{mean}}A_{\text{slope}}$ & \ding{52} \\[1ex]
    SBSR-L0~\cite{2006hugueneyadapt} & means, change-points & \ding{52} & \ding{52} & \ding{56} & $w \ceil{\log_2(A_\text{clusters})}$ & $2Nn_\text{bits}A_\text{clusters}$ & \ding{52} \\[1ex]    
    GASAX~\cite{2012fuadgasax} & means & \ding{56} & \ding{52} & \ding{52} & $w \ceil{\log_2(A_\text{mean})}$ & $n_\text{bits} A_{\text{mean}}$ & \ding{52} \\[1ex]
    aSAX~\cite{2010phamhotsax} & means & \ding{56} & \ding{52} & \ding{52} & $w \ceil{\log_2(A_\text{clusters})}$ & $n_\text{bits}A_\text{clusters}$ & \ding{52} \\[1ex]
    Sant’Anna and Wickström~\cite{2010santannaclust} & ad hoc features & \ding{52} & \ding{52} & \ding{56} & $w \ceil{\log_2(A_\text{clusters})}$ & $5 N n_\text{bits}A_\text{clusters}$ & \ding{56} \\[1ex]
    SFA~\cite{2012_schafer_sfa} & Fourier coefficients & N/A$^{\ddagger}$ & \ding{52} & \ding{52} & $w \ceil{\log_2(A_\text{coefficients})}$ & $n_\text{bits}A_\text{coefficients}$ & \ding{56} \\[1ex]
    ABBA~\cite{2020elsworthabba} & increments, lengths & \ding{52} & \ding{52} & \ding{56} & $w \ceil{\log_2(A_\text{clusters})}$ & $2Nn_\text{bits}A_\text{clusters}$ & \ding{56} \\[1ex]
    ASTRIDE & means, lengths$^{\S}$ & \ding{52} & \ding{52} & \ding{52} & $w \ceil{\log_2(A_\text{mean})}$ & $n_\text{bits}(A_\text{mean}+w)$ & \ding{52} \\[1ex]
    FASTRIDE & means & \ding{56} & \ding{52} & \ding{52} & $w \ceil{\log_2(A_\text{mean})}$ & $n_\text{bits}A_\text{mean}$ & \ding{52} \\
    \hline
  \end{tabular}
}
\end{table}


\subsection{Limitations of existing symbolization methods}
\label{subsec:limitations}

\subsubsection{The need for adaptive segmentation and quantization steps}

As stated above, only a few methods are adaptive, and fewer are adaptive on both the segmentation and quantization steps.
Let us understand with the SAX method why this can be an issue. Note that the same observations apply to SAX-like variants.

First, uniform segmentation has flaws.
Most signals found in practice contain salient events that are crucial to perform tasks such as classification.
However, uniform segmentation does not detect these events and neglects the phenomena of interest.
As can be seen on the SAX representation of Figure~\ref{fig:symbolization_example}, the two peaks around timestamps $280$ and $330$ are not detected.
Uniform segmentation does not depend on the specific signal or data set at hand, but only on the input word length $w$.
Moreover, because of it, SAX is restricted to input signals of equal length, while real-world signals are often of varying lengths~\cite{Tan2019TimeSC}.

\begin{figure}[hbtp]
  \centering
  \includegraphics[width=0.85\linewidth]{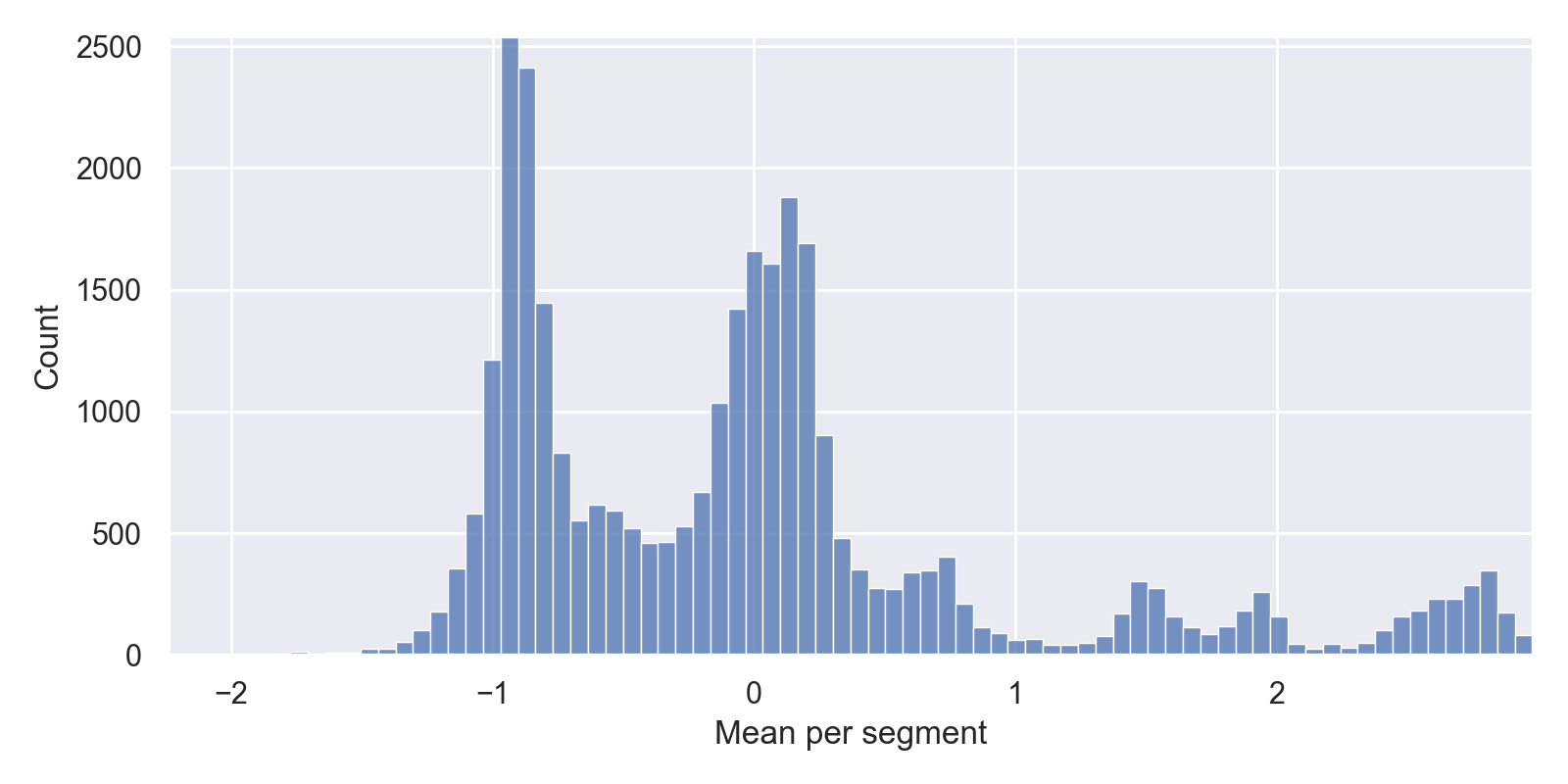}
  
  \caption{Example of histogram of the Strawberry data set (UCR Time Series Classification Archive) whose signals are of length $n=235$ and for which the word length is set to $w=32$. The obtained p-value for D'Agostino's $K^2$ normality test is $0$: the means per segment do not come from a normal distribution.}
  \label{fig:gaussian_assumption}
\end{figure}

As for quantization, the Gaussian assumption of SAX can be inappropriate for some data sets.
SAX considers that the symbols obtained after quantization will be equiprobable because all normalized time series follow a Gaussian distribution.
While normalized time series that are independent and identically distributed do tend to follow a Gaussian distribution, this in not the case for the means per segment~\cite{2015butlergauss}.
To illustrate this point, we computed the means per segment for a data set from the UCR Time Series Classification Archive~\cite{UCRArchive2018} and their histogram is displayed on Figure~\ref{fig:gaussian_assumption}. As observed, the means per segment do not seem to follow a Gaussian distribution.
We also performed the D'Agostino's $K^2$ normality test, whose null hypothesis is that the sample comes from a normal distribution.
This test rejects the Gaussian assumption at the risk level $\alpha=5\%$.
In total, we computed the mean per segment for each of 86 univariate equal-size data sets from the UCR Time Series Classification Archive (that will be taken into account in the experiments of Section~\ref{sec:expresults}).
All data sets reject the normal distribution hypothesis at the risk level $\alpha=5\%$.
Note that the authors of SAX~\cite{2007linexpsax} emphasize that, if the normality assumption is not satisfied, the algorithm is less efficient but still correct due to the lower-bounding property~\eqref{eq:lb}.

\subsubsection{The need for a similarity measure on symbolic sequences}

As seen in Table~\ref{tab:symb_summary}, some symbolic representations do not provide a similarity measure. Thus, they cannot be used directly for tasks such as classification or clustering.
As stated in Section~\ref{subsec:distsymb}, most symbolization methods that have a similarity measure are based on MINDIST from SAX.
An issue of MINDIST is that it considers adjacent symbols to be equal. For example, the MINDIST measure between adjacent symbols \texttt{a} and \texttt{b} is null, hence symbols that are actually different will be considered equal by MINDIST, which can lead to misclassification of signals.
Moreover, MINDIST is based on a Gaussian assumption.
As a result, the MINDIST distance does not adapt to the signal.
In addition, MINDIST is restricted to symbolic sequences of equal lengths.
For example, MINDIST cannot be applied to the ABBA symbolic sequences.

\subsubsection{The need for a shared dictionary of symbols across the signals of a data set}
\label{subsubsec:shared_dictionary}

As stated in the introduction, one of the goals of symbolization is to reduce the memory usage of the data.
However, only a few papers mention that, in order to reconstruct a data set of $N$ symbolic sequences, one needs to store the $N$ symbolic sequences, but also the dictionary of $A$ symbols for each signal.
Denote by $n_\text{bits}$ the number of bits needed to store a real value.
A symbolic sequence with one symbol per segment requires $w\log_2(A)$ bits, resulting in $Nw\log_2(A)$ bits for $N$ symbolic sequences.
A dictionary of symbols with one real value per symbol needs $n_\text{bits}A$ per signal, resulting in $Nn_\text{bits}A$ bits for $N$ symbolic sequences if the dictionary of symbols is not shared across signals.

Let us consider the SAX and the ABBA methods. SAX carries a shared dictionary of symbols across signals, while ABBA does not.
For SAX, $Nw\log_2(A)+n_\text{bits}A$ bits are needed to reconstruct a data set of $N$ symbolic sequences.
For the dictionary of symbols of ABBA, each symbol is a cluster center, and each cluster center has two real values: the length and the increment.
Hence, each symbol requires $2n_\text{bits}$ bits in memory. But, contrary to SAX, ABBA needs to store one dictionary of symbols per signal.
As a result, we need $Nw\log_2(A) + 2n_\text{bits}NA$ bits for the whole data set.
For the Meat data set with $N=120$, $w=10$, $A=9$, and $n_\text{bits}=64$ bits, the memory usage to encode the whole data set is $4{,}380$ bits for SAX, and $142{,}044$ bits for ABBA, thus 32 times more.
For ABBA, encoding the dictionary of symbols costs 36 times more than encoding the symbolic sequences.
As a result, ABBA requires much more memory usage than SAX because it is adaptive and its dictionary of symbols is not shared.
The memory usage to reconstruct a data set of symbolic sequences for more methods are given Table~\ref{tab:symb_summary}.

\subsection{Contributions}

To the best of our knowledge, the ASTRIDE method to be presented in Section~\ref{sec:descriptionastride} is the only symbolic representation offering adaptive segmentation and quantization, a shared dictionary of symbols as well as a compatible similarity measure and a reconstruction procedure.
Altogether, ASTRIDE circumvents the limitations of the methods described in Section~\ref{subsec:limitations}.

Instead of using uniform segmentation, ASTRIDE performs adaptive segmentation, a.k.a. change-point detection~\cite{ruptures}, in order to capture salient events.
More precisely, we detect changes in the mean, where the number of changes is set by the user.
Moreover, ASTRIDE does not rely on the Gaussian assumption for the quantization: this step is adaptive on the signals at hand.
Consequently, ASTRIDE does not require any assumption on the distribution of the data.

We also introduce Dynamic General Edit Distance (D-GED), a new similarity measure on symbolic representations which is based on the general edit distance.
Moreover, unlike MINDIST, the symbolic sequences are not required to be of equal lengths.

Adaptive segmentation and quantization are learned at the level of the data set of signals: the change-points as well as the quantiles (for the quantization) are estimated using all signals in the data set.
ASTRIDE's dictionary of symbols is the same for all signals (unlike ABBA), and is thus memory-efficient.

\section{The ASTRIDE method}\label{sec:descriptionastride}

ASTRIDE (Adaptive Symbolization Time seRIes DatabasEs) is a novel symbolic representation for data sets of signals.
It is an offline method inputting univariate signals that are required to be of equal size.
There are two parameters to be set by the user: the word length $w$ and the alphabet size $A$.
ASTRIDE comes with a new similarity measure on symbolic sequences: D-GED (Dynamic General Edit Distance).
After describing ASTRIDE and D-GED, we introduce FASTRIDE (Fast ASTRIDE), an accelerated version of ASTRIDE.

\subsection{ASTRIDE segmentation step}

As a preprocessing step, all times series in the data set are centered and scaled to unit variance.
Then, the $N$ signals of length $n$ are segmented.
To that end, all signals are stacked, producing a single multivariate signal of length $n$ and dimension $N$.
ASTRIDE applies multivariate change-points detection with a fixed number of segments on this high-dimensional signal.
When $w$ segments are chosen, the segmentation provides $w-1$ change-points that are the same for each univariate signal.
Since the change-points are common to all (univariate) signals, this allows ASTRIDE to be memory-efficient.
The lengths of each resulting symbolic sequence are the same (equal to $w$).
For a given multivariate signal $\bm{y} = (y_1, \ldots, y_n)$ with $n$ samples, change-point detection finds the $w-1$ unknown instants $t_1^* < t_2^* < \ldots < t_{w-1}^*$ where some characteristics (here, the mean) of $\bm{y}$ change abruptly.
A recent review of such methods is given in~\cite{ruptures}.
In the context of ASTRIDE, the number of changes $w-1$ is chosen by the user: it is the desired number of regimes, meaning the length of the resulting symbolic sequences.
The change-point algorithm estimates $\hat{t}_1, \ldots, \hat{t}_{w-1}$ which are the minimizers of a discrete optimization problem:
\begin{equation}\label{eq:ruptures}
    \left(\hat{t}_1, \ldots, \hat{t}_{w-1}\right) = \argmin_{(w, t_1, \ldots, t_{w-1})} \sum_{k=0}^{w+1} \sum_{t=t_k}^{t_{k+1}-1} \|y_t - \bar{y}_{t_k:t_{k+1}}\|^2 
\end{equation}
where $\bar{y}_{t_k:t_{k+1}}$ is the empirical mean of $\{y_{t_k},\dots,y_{t_{k+1}-1}\}$.
By convention, $t_0=0$ and $t_{w}=n$.
Formulation~\eqref{eq:ruptures} seeks to reduce the error between the original signal and the best piecewise constant approximation.
This problem is solved using dynamic programming which has a time complexity of $O\left(Nwn^2\right)$ where $N$ is the number of signals in the data set.

\begin{figure}[hbtp]
    \centering
    \includegraphics[width=\linewidth]{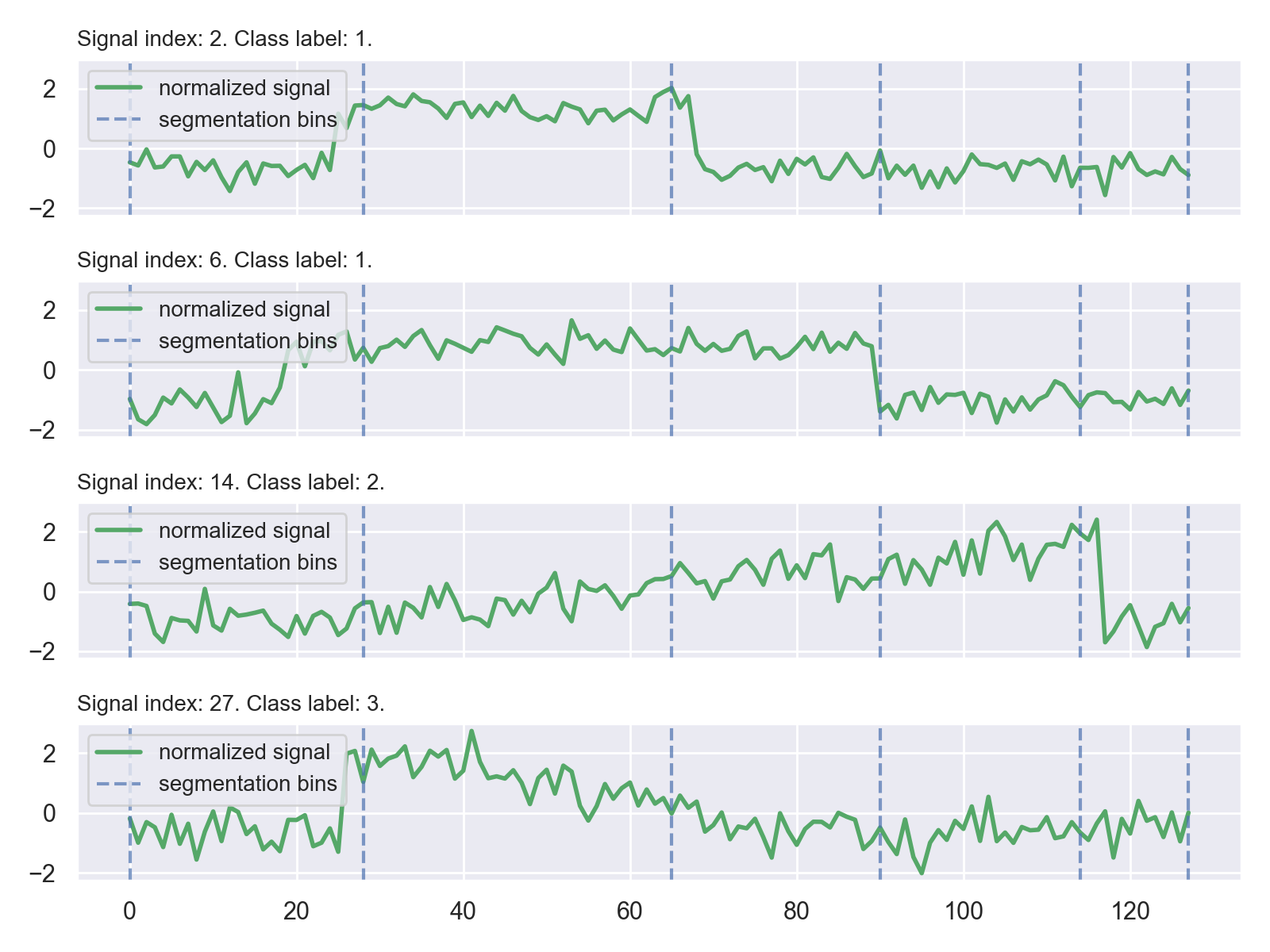}
    \caption{Multivariate change-point detection on (univariate) signals from the CBF data set (UCR Time Series Classification Archive).
    Here, $n=128$ and $w=5$. The change-points are obtained from the whole training set, but only a few signals are displayed.}
    \label{fig:multivariate_ruptures_CBF}
\end{figure}

Figure~\ref{fig:symbolization_example} displays an example of an ASTRIDE representation of a signal, along with the SAX representation (for the same parameters $w$ and $A$). 
Visually, compared to uniform segmentation, adaptive segmentation leads to more meaningful segments.
For example, it detects that one segment is sufficient to approximate the signal from timestamp $0$ to $250$, and that there is a peak around timestamp $280$ and another one around timestamp $330$.
It shows the importance of our adaptive segmentation scheme.
Figure~\ref{fig:multivariate_ruptures_CBF} depicts how the multivariate change-point detection works. The algorithm tries to find the abrupt changes in mean that are common to most (univariate) signals in the data set.

\subsection{ASTRIDE adaptive quantization step}

After segmentation, the means of all segments are computed and grouped into bins based on the empirical quantiles.
Each segment is then symbolized by the bin it belongs to.
This quantization step is similar to the MCB (Multiple Coefficient Binning) procedure of SFA~\cite{2012_schafer_sfa}.
Since the segments found during the segmentation step correspond to mean-shifts, it is reasonable to represent each segment by its mean value.
The $A-1$ quantiles are calculated on the means of all segments of all signals in the data set, leading to $A$ symbols.
The time complexity of the quantization step (computing the means, the quantiles, and applying the binning) is $O\left(Nw\right)$, where $N$ is the number of signals in the data set.
By design, all symbols are equiprobable.
Figure~\ref{fig:symbolization_example} shows an example of an ASTRIDE representation. Compared to SAX, the bins of ASTRIDE represent the quantiles of the means per segment and are quite different from the ones of SAX.
Recall that ASTRIDE is fitted on the whole training set (and not on the displayed signal only).

\subsection{The D-GED similarity measure}
\label{subsec:dged}

We introduce Dynamic General Edit Distance (D-GED), a novel similarity measure on symbolic representations.
D-GED is compatible with symbolic sequences of equal or varying lengths.
The similarity measure D-GED is based on the general edit distance~\cite{2001navarrostring} that is described in Section~\ref{subsec:distsymb}.
D-GED sets the operation costs of the general edit distance so that they incorporate the distance between individual symbols as follows:
\begin{itemize}
    \item The substitution cost $\operatorname{sub}(\texttt{a},\texttt{b})$ for individual symbols \texttt{a} and \texttt{b} is the Euclidean distance between the mean $\mu_\texttt{a}$ of the mean values attributed to symbol \texttt{a} and the mean $\mu_\texttt{b}$ of the mean values attributed to symbol \texttt{b}
\begin{equation}
    \label{eq:sub}
    \operatorname{sub}(\texttt{a},\texttt{b}) = \left\|\mu_\texttt{a} - \mu_\texttt{b}\right\|_2.
\end{equation}
    \item For all characters, the insertion and deletion costs are set to $\operatorname{sub}_{\max}$, where $\operatorname{sub}_{\max}$ is the maximum value of the modified substitute costs given in~\eqref{eq:sub}.
\end{itemize}
For the substitution cost, the intuition is that if symbols \texttt{a} and \texttt{b} are "very different", then the difference between $\mu_\texttt{a}$ and $\mu_\texttt{b}$ will be wider, and substituting them will have a larger cost in D-GED.
By setting the insertion and deletion costs to $\operatorname{sub}_{\max}$, D-GED favors substitutions over insertions and deletions.
The worst-case complexity to compute the D-GED similarity measure of two symbolic sequences of lengths $w_1$ and $w_2$ is $O(w_1w_2)$.

D-GED is not applied directly on the symbolic representation but on a replicated version.
Indeed, when a method uses a non-uniform segmentation, the segments can have different lengths.
Without taking into account the varying segment sizes, D-GED would compare (substitute or delete/insert) symbols corresponding to segments of different lengths.
To prevent ASTRIDE from losing this information, we propose the following procedure.
Denote by $\ell_1, \ldots, \ell_w$ the segment lengths obtained with our adaptive segmentation.
By design, they are the same for all signals in the data set.
Each segment length is divided by the minimum of all segments lengths and rounded to its nearest integer to obtain the \emph{normalized segment lengths} $\widehat{\ell}_1, \ldots, \widehat{\ell}_w$.
Then, the symbolic sequences are modified by replicating the symbol of the first segment $\widehat{\ell}_1$ times, then the symbol of the second segment $\widehat{\ell}_2$ times, etc.
Finally, the D-GED measure between these replicated symbolic sequences is computed.
As an example, consider the symbolic sequence from ASTRIDE depicted in Figure~\ref{fig:symbolization_example}.
The symbolic sequence without incorporating information about the segment lengths is \texttt{1230}.
The segment lengths are $(266, 47, 40, 95)$ before normalization (more details on the signal are given in Table~\ref{tab:quantif_length_regimes}).
The smallest segment has 40 samples and, as a result, the normalized segment lengths are $(7, 1, 1, 2)$.
\begin{table}[hbtp]
    \centering
    \begin{minipage}{\linewidth}
    \caption{Details of the ASTRIDE representation of the signal displayed on Figure~\ref{fig:symbolization_example}. The parameters of ASTRIDE are $w=4$ and $A=4$. (The quantized mean feature is described in Section~\ref{subsec:astride_reconstruction}.)}
    \label{tab:quantif_length_regimes}
    \centering
    \medbreak
    \begin{tabular}{@{}rrrrrr@{}}
    \hline
     segment start & mean & symbol & quantized mean & length & normalized length \\
    \hline
    0 & $-0.17$ & \texttt{1} & $-0.16$  & 266 & 7 \\
    266 & $1.11$ & \texttt{2} & $1.05$ & 47 & 1 \\
    313 & $2.34$ & \texttt{3} & $2.38$ & 40 & 1 \\
    353 & $-1.07$ & \texttt{0} & $-1.06$ & 95 & 2 \\
    \hline
    \end{tabular}
    \end{minipage}
\end{table}
The replicated symbolic sequence based on the normalized segment lengths is
\begin{equation}
    \underbrace{\text{\texttt{1111111}}}_{7\ \text{times}} \underbrace{\text{\texttt{2}}}_{\text{once}} \underbrace{\text{\texttt{3}}}_{\text{once}} \underbrace{\text{\texttt{00}}}_{\text{twice}}
    \label{eq:partial_replication}
\end{equation}
which is of total length 11.

\subsection{Reconstruction of ASTRIDE symbolic sequences}
\label{subsec:astride_reconstruction}

\begin{figure}[hbtp]
    \centering
    \includegraphics[width=\linewidth]{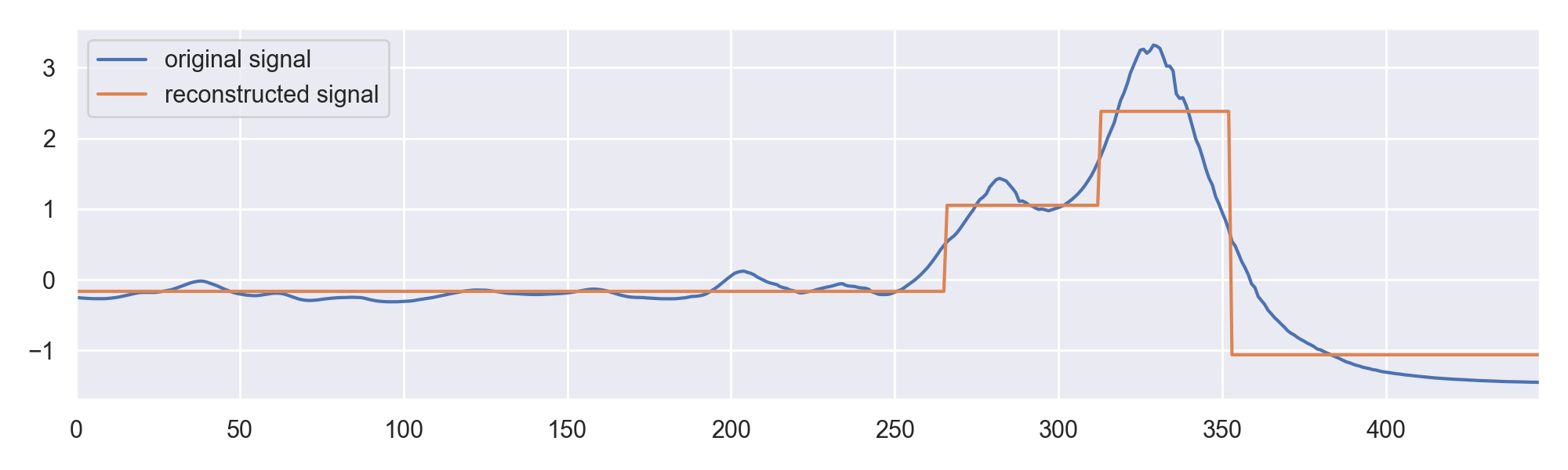}
    \caption{Example of reconstruction by ASTRIDE of the \texttt{1230} symbolic (same original signal as in Figure~\ref{fig:symbolization_example}). Here, $w=4$ and $A=4$. Note that ASTRIDE is fitted on the whole training set (and not on the displayed signal only).}
    \label{fig:reconstruction}
\end{figure}

In ASTRIDE, a signal is reconstructed from its symbolic sequence as follows.
Each symbol of the symbolic representation is replicated $\ell_k$ times, where $\ell_k$ is the length of the associated segment.
The length of the reconstructed signal is the same as the original one.
Then, each symbol is replaced by the average of all segment means that belong to the associated bin (the quantized mean).
This resulting real-valued signal is the reconstruction by ASTRIDE.
As an example, consider the signal shown in Figure~\ref{fig:symbolization_example} whose symbolic representation is \texttt{1230}.
Details about the segment lengths, symbols, and quantized mean are given in Table~\ref{tab:quantif_length_regimes}.
First, the  symbols \texttt{1}, \texttt{2}, \texttt{3} and \texttt{0} are replicated 266, 47, 40 and 95 times respectively.
Second, each symbol is mapped to its quantized mean, going from a symbolic signal to a real-valued signal:
\begin{equation}
    \underbrace{\fbox{-0.16}\ldots\fbox{-0.16}}_{266\ \text{times}} \underbrace{\fbox{1.05}\ldots\fbox{1.05}}_{47\ \text{times}} \underbrace{\fbox{2.38}\ldots\fbox{2.38}}_{40\ \text{times}} \underbrace{\fbox{-1.06}\ldots\fbox{-1.06}}_{95\ \text{times}}
    \label{eq:reconstructed_signal}
\end{equation}
The real-valued signal displayed in Formula~\eqref{eq:reconstructed_signal} is the reconstruction of the \texttt{1230} symbolic sequence.
A reconstructed signal from ASTRIDE is a piecewise constant signal, as displayed in Figure~\ref{fig:reconstruction}. 
Notice how the reconstructed signal in Figure~\ref{fig:reconstruction} is different from the mean per segment representation in Figure~\ref{fig:symbolization_example}, as the quantized mean is used, and not the mean.

The memory cost of ASTRIDE is easily derived.
To reconstruct $N$ symbolic sequences from the ASTRIDE representation with $w$ segments and $A$ symbols, one needs to store $Nw\log_2(A)$ bits for the symbolic sequences.
For the shared dictionary of symbols, one needs $n_\text{bits}A$ bits.
In addition to storing the symbolic sequences with their shared dictionary of symbols, one also needs to store the $w$ segment lengths that are the same for all symbolic sequences, resulting in $n_\text{bits}w$ bits for the whole data set. In total, $Nw\log_2(A) + (w+A)n_\text{bits}$ are required.
To illustrate, let us take the example in Section~\ref{subsubsec:shared_dictionary}: the Meat data set with $N=120$, $w=10$, $A=9$, and $n_\text{bits}=64$ bits.
For ASTRIDE, the memory usage to encode the whole data set is $5{,}020$ bits.
Recall that the memory usage is $4{,}380$ bits for SAX, and $142{,}044$ bits for ABBA.

\subsection{The FASTRIDE method}

FASTRIDE (Fast ASTRIDE) is an accelerated variant of ASTRIDE.
For the symbolization procedure, the only difference is its segmentation step which is uniform, like SAX.
The reconstruction of FASTRIDE is performed in the same way as ASTRIDE.
For the similarity measure of FASTRIDE, we use D-GED but there is no need to replicate the symbolic sequences, as the segment lengths are equal due to the uniform segmentation.
FASTRIDE is computationally faster than ASTRIDE because FASTRIDE skips the adaptive segmentation step, and the input symbolic sequences of the general edit distance are not replicated, thus are shorter.

\section{Experimental results}\label{sec:expresults}

We compare ASTRIDE and FASTRIDE to several popular symbolic representations (SAX, 1d-SAX, SFA, ABBA) on a classification task and a reconstruction task.
We show that ASTRIDE and FASTRIDE constitute the best compromises to address both these tasks. Indeed, as will be discussed, some of these methods can be used only on one task; in particular, SFA and ABBA do not possess a similarity measure and, therefore, cannot be used as such for classification.

The adaptive segmentation step of ASTRIDE is implemented with the \texttt{ruptures} Python package~\cite{ruptures}.
The general edit distance in D-GED uses the \texttt{weighted-levenshtein} Python package~\cite{weighted-levenshtein}.
SAX and 1d-SAX are implemented in the \texttt{tslearn} Python package~\cite{tslearn}.
SFA is implemented from scratch.
ABBA is taken from the authors' GitHub repository\footnote{\url{https://github.com/nla-group/ABBA}}.
A Python implementation of ASTRIDE and FASTRIDE, along with codes to reproduce the figures and scores in this paper, can be found in a GitHub repository\footnote{ \url{https://github.com/sylvaincom/astride}}.

\subsection{Classification task}
\label{subsec:bench_classif}

We first investigate the performances of our approaches on a classification task.

\subsubsection{Experimental setup}
\label{subsubsec:bench_classif_setup}

\paragraph{Data mining task and competitors}
Our methods ASTRIDE and FASTRIDE are compared to SAX and 1d-SAX.
One-Nearest Neighbor (1-NN) classification is used to compare the quality of both the symbolizations and the  similarity measures, as often done in the literature~\cite{bagnall2017great}.
For ASTRIDE and FASTRIDE, the change-points and the quantization bins are learned on the training set.

Our comparison is limited to classification techniques based on symbolizations, since our objective is to evaluate the relevance of this step itself and not to achieve state-of-the-art performance on time series classification.
Hence, we exclude classifiers that are built on top of symbolic representations, namely bag-of-words and ensemble-based algorithms. In particular, SAX-VSM (SAX and Vector Space Model)~\cite{2013_senin_saxvsm}, BOSS (Bag-of-SFA-Symbols)~\cite{2014_schafer_boss}, Mr-SEQL (Multiple symbolic representations SEQuence Learner)~\cite{Nguyen2019InterpretableTS}, WEASEL (Word ExtrAction for time SEries cLassification)~\cite{Schfer2017FastAA}, and TDE (Temporal Dictionary Ensemble)~\cite{Middlehurst2020TheTD} are out of the scope on this paper.
More details on these techniques can be found in~\cite{bagnall2017great, Ruiz2021TheGM, Agarwal2021RankingBA}.

\paragraph{Hyperparameters}
The hyperparameters for all methods are:
\begin{itemize}
    \item the word length $w$ in $\{5, 10, 15, 20, 25\}$
    \item the alphabet size $A$ in $\{4, 9, 16, 25\}$.
\end{itemize}
For 1d-SAX, $A \in \{4, 9, 16, 25\}$ corresponds to $(A_\text{mean}, A_\text{slope}) \in \{(2, 2), (3, 3), (4, 4), (5, 5)\}$.

\paragraph{Evaluation of the task}
The evaluation metric for the classification is the test accuracy: percentage of correctly classified signals.

\paragraph{Data sets}
Since SAX and other methods can be applied only to univariate and equal-size times series and we choose our signals to be of length at least 100 (as in the ABBA paper~\cite{2020elsworthabba}), the scope of our comparisons is restricted to 86 data sets of the UCR Times Series Classification Archive~\cite{UCRArchive2018}.
The data sets are both real-world and synthetic, and come with a default train / test split which is the one used in this paper.
Our experiments were launched on a total of 66,827,003 samples across all signals of all data sets.
Table~\ref{tab:ucr} recaps some key figures of the considered data sets.
While averaging accuracies over different data sets, with different sizes and challenges, has some flaws, it is the best compromise to obtain a global key figure to assess the quality of a classifier on the UCR Time Series Classification Archive.
\begin{table}[tp]
    \centering
    \begin{minipage}{0.8\linewidth}
    \caption{Presentation of the 86 univariate equal-size data sets from the UCR Time Series Classification Archive considered in our classification experiment.}
    \label{tab:ucr}
    \centering
    \medbreak
    \begin{tabular}{@{}lrrr@{}}
    \hline
    {} &  No. of signals &  Length &  No. of classes \\
    \hline
    mean & 1,357 & 644 & 10 \\
    min & 40 & 128 & 2 \\
    50\% & 687 & 456 & 4 \\
    max & 9,236 & 2,844 & 60 \\
    \hline
    \end{tabular}
    \end{minipage}
\end{table}

\subsubsection{Results}

\begin{figure}[hbtp]
    \centering
    \includegraphics[width=0.93\linewidth]{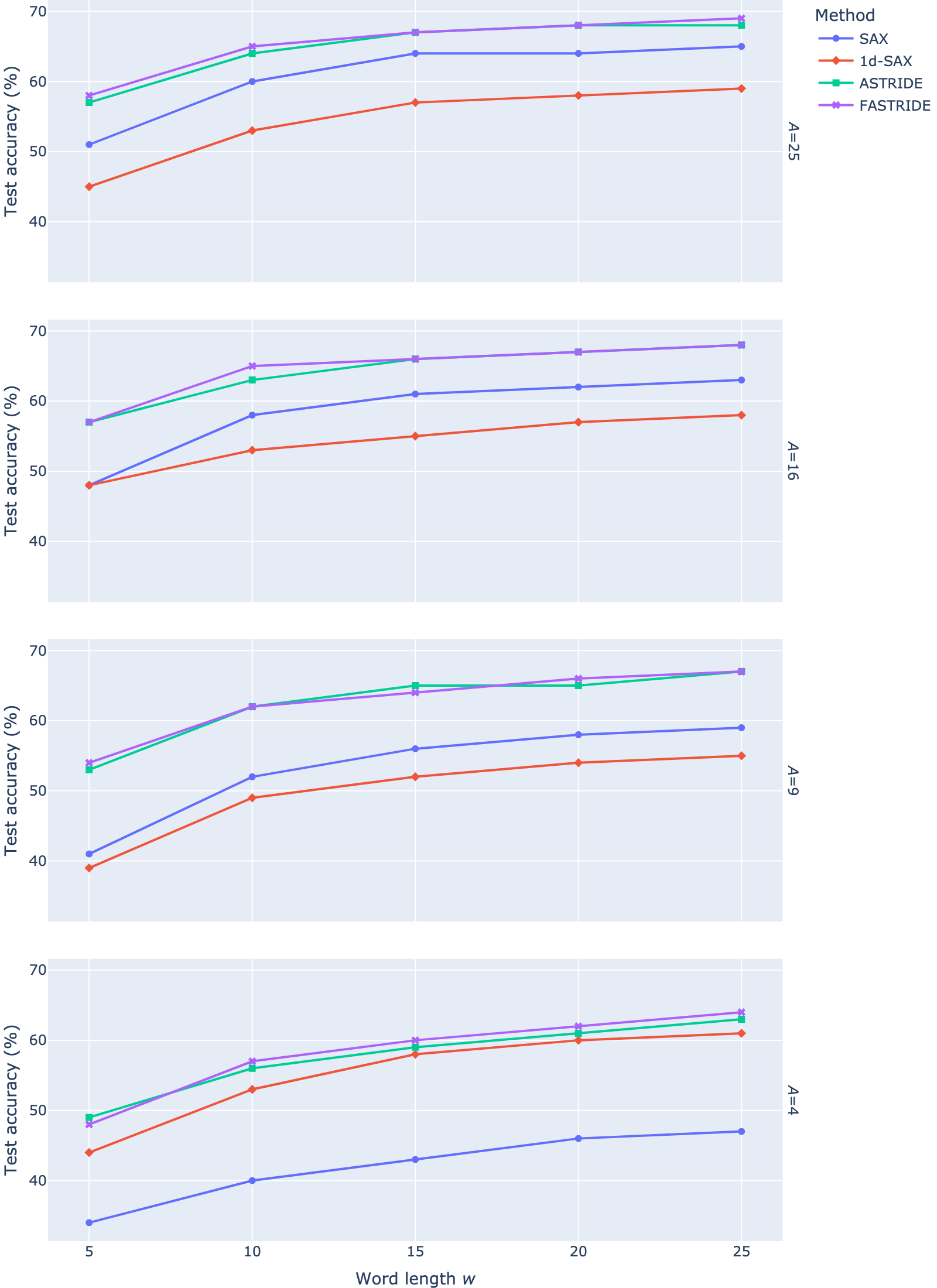}
    
    \caption{Accuracy of SAX, 1d-SAX, ASTRIDE, and FASTRIDE on the classification task versus the word length $w$, for several values of alphabet size $A$, averaged on 86 data sets from the UCR Time Series Classification Archive.}
    \label{fig:classif_benchmark}
\end{figure}

Figure~\ref{fig:classif_benchmark} displays the accuracy scores as a function of the word length $w$ averaged over the selected data sets, for several methods, and for different values of alphabet size $A$.
For each method and each alphabet size $A$, plotting the accuracy in relation to $w$ tells us which method provides the best representation: the larger the accuracy for a given $w$, the better the symbolic representation.

Our results show that ASTRIDE and FASTRIDE perform better than both SAX and 1d-SAX on the classification task.
Indeed, for each alphabet size $A$ and each word length $w$, ASTRIDE and FASTRIDE have a higher accuracy than both SAX and 1d-SAX. This shows that the proposed adaptive symbolization process, combined with the D-GED similarity measure, is relevant in this classification context.

\paragraph{Influence of the parameters}
We observe that, for all methods, the accuracy increases as the word length $w$ increases.
This was expected as the symbolization becomes more precise as each signal is represented with more bits.
Interestingly, ASTRIDE and FASTRIDE achieve reasonable classification results even for a very small number of segments.
For example, when $w=5$ and $A=16$, ASTRIDE has a score of $57\%$, while SAX and 1d-SAX have a score of $48\%$.
This confirms the fact that using a more adaptive representation better captures the phenomenon observed in the signals, and thus better compresses the information.

For all methods except 1d-SAX, when the alphabet size $A$ increases, the classification scores improve.
Yet, ASTRIDE and FASTRIDE are less sensitive to the value of $A$.
For $w=20$, the accuracy of ASTRIDE is of $61\%$, $65\%$, $67\%$, and $68\%$, for $A=4$, $A=9$, $A=16$, and $A=25$ respectively.
On the contrary, SAX seems to be very sensitive to the value of $A$: for $w=20$, it reaches 45\% for $A=4$ and 62\% for $A=16$. Moreover, the performance of SAX is worse than 1d-SAX for small values of $A$, and is slightly better than 1d-SAX (and even largely surpasses it) for large values of $A$.

\paragraph{Importance of the adaptive approaches}
According to Figure~\ref{fig:classif_benchmark}, FASTRIDE achieves results similar to those of ASTRIDE, which suggests that the adaptive segmentation does not increase performances. As will be seen in the next section, the relevance of the segmentation phase is more acute in the reconstruction task.

The importance of the adaptive quantization process based on quantiles can been assessed by comparing the performances of FASTRIDE to those of SAX, which uses quantization bins based on the standard normal distribution instead of an empirical distribution.
Based of the results, it appears that using the quantiles to build both the symbolization and the similarity measure allows us to adapt it to the data set of interest, and to detect variations between signals that are not captured by the fixed MINDIST costs.

We also note that both FASTRIDE and ASTRIDE benefit from the newly introduced D-GED distance, which offers nice performances on this classification task.

\subsection{Reconstruction task}
\label{subsec:bench_reconstruction}

In this section, we investigate the performances of our approaches on a reconstruction task.

\subsubsection{Experimental setup}
\label{subsubsec:bench_reconstruction_setup}

\paragraph{Data mining task and competitors}
Our ASTRIDE and FASTRIDE representations, SAX, 1d-SAX, SFA, and ABBA are compared on a reconstruction task.
Except for ABBA, the papers about SAX, 1d-SAX, and SFA do not tackle signal reconstruction. However, it is easy to infer a reconstruction procedure for these methods.
For SAX and 1d-SAX, the sample values on each segment of the reconstructed signal are based on the Gaussian bins.
For SFA, the reconstructed signal is the Fourier reconstruction based on the quantized Fourier coefficients.

\paragraph{Hyperparameters}
The alphabet size is fixed to $A=9$ for all methods, and the scaling parameter of ABBA is set to $scl=1$, as is done in~\cite{2020elsworthabba}\footnote{Note that we obtain similar reconstruction error results with $scl=0$.}.
The word length $w$ results from the choice of the tolerance $tol$ in ABBA.
For fair comparison, this value of $w$ is determined by running ABBA with a fixed value of $tol$.
To compare the different approaches, we apply a protocol inspired from the ABBA paper~\cite{2020elsworthabba}.
For each signal, ABBA is first run with a low tolerance $tol = 0.05$ and it returns the number of segments $w$ to approximate the original signal at tolerance $tol$.
If $w \geq n \tau_t$, where $n$ is the length of the original signal and $\tau_t$ is a target memory usage ratio, we successively increase $tol$ by $0.05$ and rerun ABBA until $w \leq n \tau_t$.
This last value is denoted by $w_e$.
As in~\cite{2020elsworthabba}, we exclude all signals leading to $w_e < 9$, because we choose $A=9$.

We run this protocol for $\tau_t \in \{5\%, 6.7\%, 10\%, 16.7\%, 20\%, 25\%, 33.3\%\}$.
Unlike ABBA, which works signal by signal, SFA, ASTRIDE, and FASTRIDE work on a whole data set, so their input word length $w$ is the same for all signals in the data set.
For the latter methods, we set $w$ equal to the average $\overline{w}_e$ of the $w_e$'s obtained for each signal by ABBA.
As a result, each data set and each value of $\tau_t$ is associated with a word length $\overline{w}_e$.

In most cases, the empirical memory usage ratio $\tau_e = \overline{w}_e/n$ is smaller than the target memory usage ratio $\tau_t$. The values of $\tau_t$ and their corresponding $\tau_e$ (averaged on all signals irrespective of their data set) are displayed in Table~\ref{tab:denom}.
In the protocol, if it is not possible to compress a signal at a given $tol$, $\tau_t$, and $A$, then the whole data set is excluded from our benchmark, which explains why there are different numbers of compatible data sets in Table~\ref{tab:denom}.

\paragraph{Evaluation of the tasks}
The evaluation metrics of the reconstruction task are the Euclidean and DTW (Dynamic Time Warping) which is robust to time-shifts.
A data set of $N$ signals is transformed into $N$ symbolic sequences, then these $N$ symbolic sequences are reconstructed.
For each signal, we compute the reconstruction error: the distance between the original signal and its reconstruction.
Recall that, for all methods, each signal is first centered and scaled to unit variance. The reconstruction error is between the scaled original signal and its reconstruction, as the normalization is important when conducting benchmarks~\cite{Keogh2002OnTN}.

Moreover, because we noticed that the SAX and 1d-SAX implementations from \texttt{tslearn (v0.5.2)}~\cite{tslearn} poorly handles the last samples of the reconstructed signals when $n$ is not divisible by $w$, the reconstruction error is computed between the truncated signals: both the original and reconstructed signals (for all methods) are truncated so that their length is $\left\lfloor n/w \right\rfloor w$.

\paragraph{Data sets}
As for the classification task, we use as input the  UCR Time Series Classification Archive.
The scope of our comparisons is restricted to equal-size data sets of length at least 100 from the UCR Times Series Classification Archive~\cite{UCRArchive2018}.
\begin{table}[hbtp]
\centering
\begin{minipage}{0.8\linewidth}
    \caption{Empirical protocol to set the word length $w$ per data set for the reconstruction benchmark, given a target memory usage ratio $\tau_t$ for ABBA.}
    \label{tab:denom}
    \centering
    \medbreak
    \begin{tabular}{@{}p{0.25\linewidth}p{0.26\linewidth}p{0.25\linewidth}@{}}
        \hline
        Target memory usage ratio $\tau_t$ (\%) & Empirical memory usage ratio $\tau_e$ (\%) & Number of compatible data sets\\
        \hline
        5 & 4.0 & 44 \\
        6.7 & 5.3 & 49 \\
        10 & 7.6 & 56 \\
        16.7 & 11.7 & 64 \\
        20 & 13.4 & 65 \\
        25 & 15.8 & 67 \\
        33.3 & 19.3 & 69 \\
        \hline
    \end{tabular}
\end{minipage}
\end{table}

\subsubsection{Results}
\label{subsubsec:results_abba}

\begin{figure}[hbtp]
    \centering
    \includegraphics[width=0.9\linewidth]{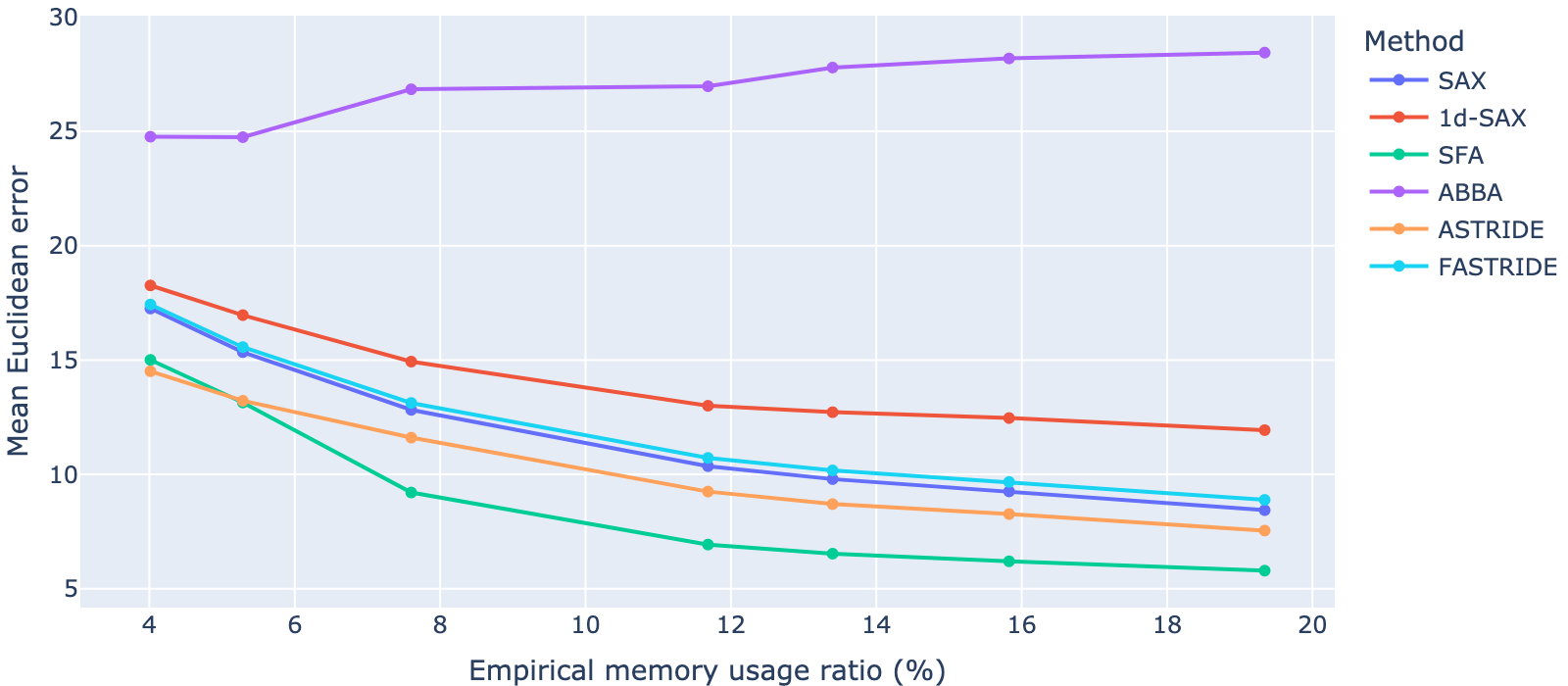}

    \bigbreak
    \includegraphics[width=0.9\linewidth]{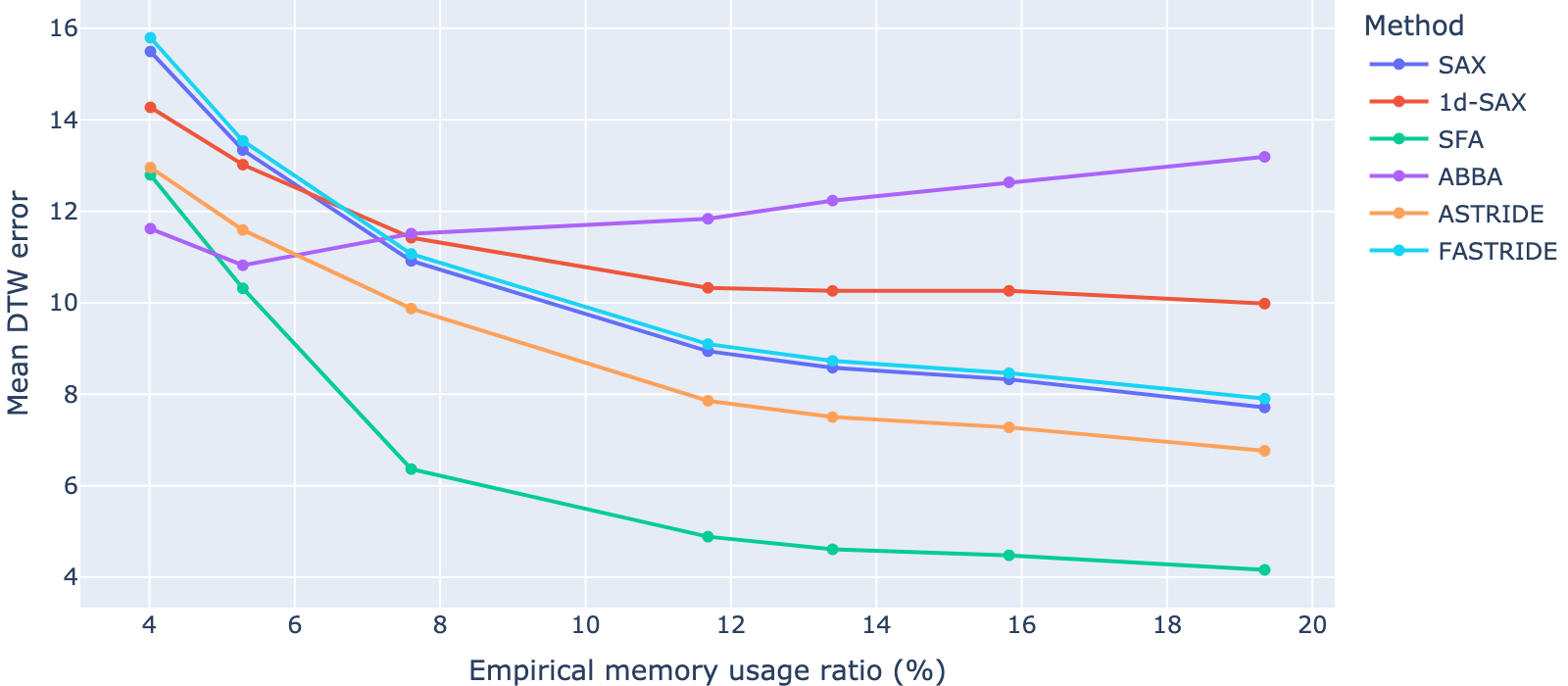}
    
    \caption{Reconstruction benchmark for several methods and several empirical memory usage ratios $\tau_e$, averaged over the signals from various data sets from the UCR Time Series Classification Archive, for $A=9$.}
    \label{fig:bench_reconstruct}
\end{figure}

Figure~\ref{fig:bench_reconstruct} displays the reconstruction error versus the memory usage ratio, averaged over the compatible data sets, for several methods, and a fixed alphabet size $A=9$. 
We observe that, for both the Euclidean and DTW errors, SFA obtains the best performances, followed by ASTRIDE.

\begin{figure}[hbtp]
    \centering
    \begin{minipage}{0.49\linewidth}
        \centering
        \includegraphics[width=0.95\linewidth]{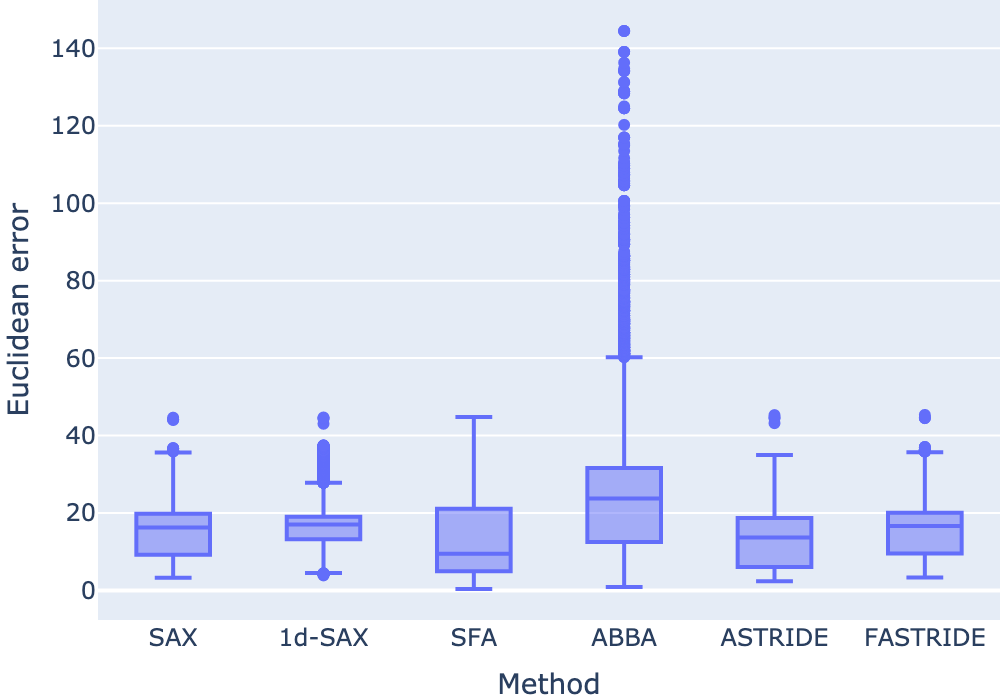}
    \end{minipage}
    \begin{minipage}{0.49\linewidth}
        \centering
        \includegraphics[width=0.95\linewidth]{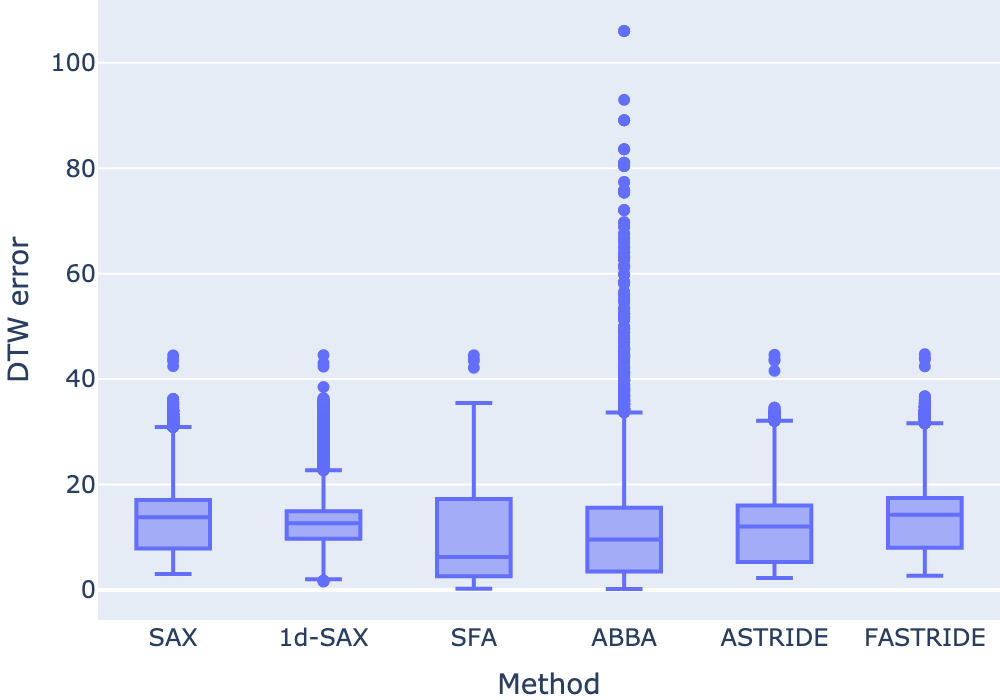}
    \end{minipage}
    
    \caption{Box-plots of the reconstruction benchmark for several methods over the signals of 49 data sets from the UCR Time Series Classification Archive, for $A=9$ and $\tau_t=6.7\%$ (leading to $\tau_e=5.3\%$).
    For both box-plots, the range of the $y$-axis is limited to the $99.99$\% quantile of the reconstruction error for visualization purposes.
    }
    \label{fig:bench_reconstruct_boxplot}
\end{figure}

To complement the use of the mean in Figure~\ref{fig:bench_reconstruct}, the box-plots in Figure~\ref{fig:bench_reconstruct_boxplot} show the spread of the reconstruction errors and the outliers, for a fixed target memory usage ratio $\tau_t=6.7\%$.
The most extreme outliers are generated by ABBA, which shows that this method is not robust.
A possible explanation is that ABBA is very sensitive to noisy signals because of its piecewise linear approximation step and its use of the quantized increments.
On the contrary, SAX, 1d-SAX, SFA, ASTRIDE, and FASTRIDE seem quite robust.

\paragraph{Influence of the memory usage ratio}
For SAX, 1d-SAX, SFA, ASTRIDE, and FASTRIDE, the reconstruction error decreases as the memory usage ratio increases. Indeed, as the memory usage ratio increases, more segments are allowed in the symbolic representations, resulting in a higher quality of the reconstruction.

For very small memory usage ratios ($\tau_e \leq 6\%$), SFA and ASTRIDE have similar performances. Moreover, when $\tau_e \leq 6\%$, according to the DTW error, ABBA performs better than ASTRIDE and FASTRIDE. However, this observation is challenged by the box-plot in Figure~\ref{fig:bench_reconstruct_boxplot}, which shows that ABBA reaches very large errors and is much less robust than ASTRIDE.
Moreover, the empirical memory usage ratio $\tau_e=\overline{w}_e/n$ in Figure~\ref{fig:bench_reconstruct} does not take into account the total memory usage: it ignores the dictionary of symbols and the fact that the reconstruction is done on a data set of signals (and not on a single signal).
As emphasized in Section~\ref{subsubsec:shared_dictionary} and Section~\ref{subsec:astride_reconstruction}, the total memory usage of ABBA is much larger than those of ASTRIDE and FASTRIDE, because ABBA does not share a dictionary of symbols across signals.

\paragraph{Importance of the adaptive approaches}
According to Figure~\ref{fig:bench_reconstruct}, ASTRIDE achieves better results than FASTRIDE, thus showing the relevance of the adaptive segmentation on the reconstruction task. Indeed, the segmentation phase allows to focus on the events of interest, which are thus correctly reconstructed. Dedicating a memory size for the fine encoding of these events thus seems to be a good strategy to compress the information contained in the signals. Adaptive quantization based on quantiles does not appear particularly useful for signal reconstruction, as FASTRIDE performs similarly to SAX.

\begin{figure}[hbtp]
    \centering
    \includegraphics[width=\linewidth]{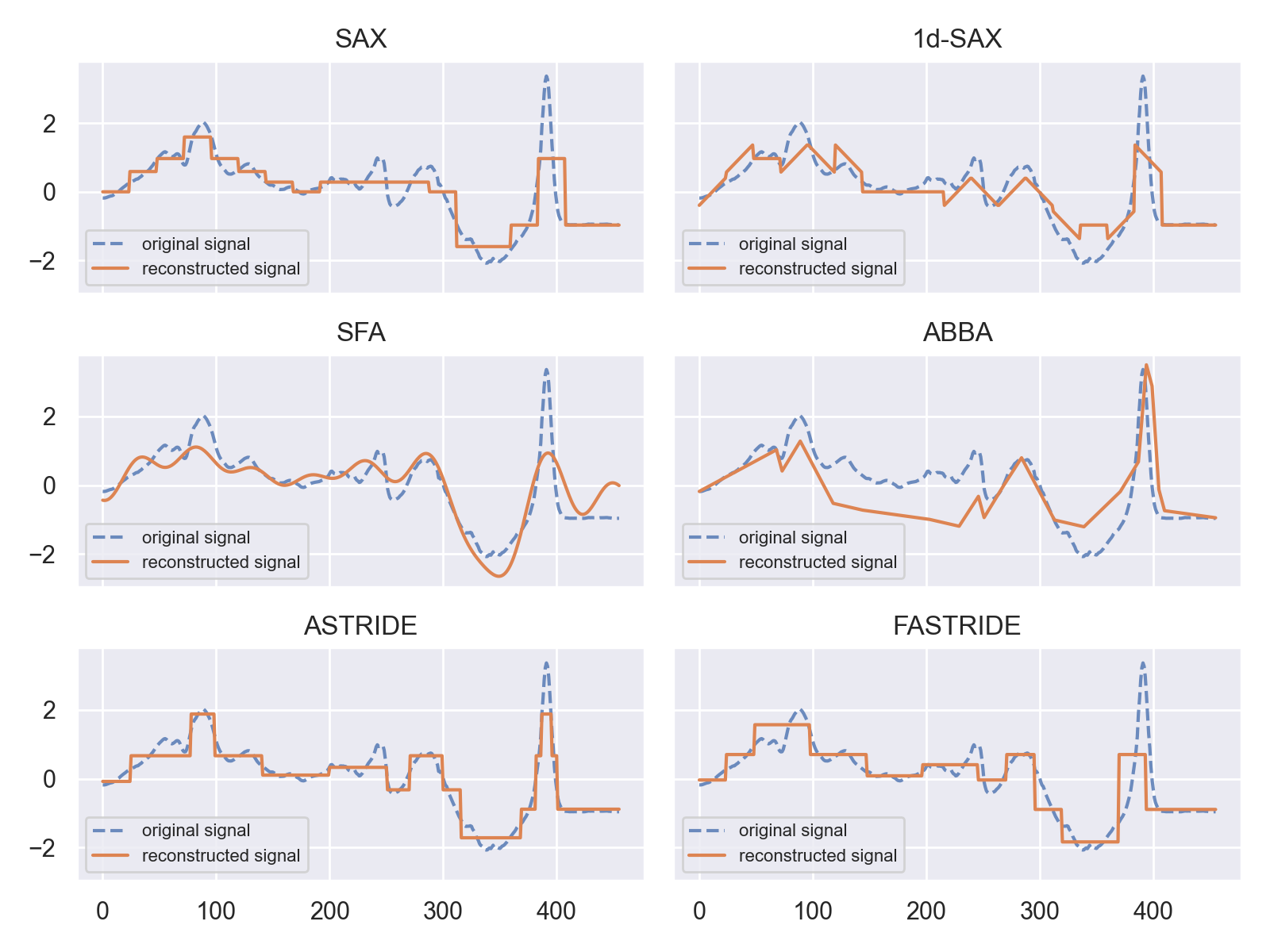}

    \caption{Example of reconstruction of a single signal from the Beef data set (UCR Time Series Classification Archive) of original length $n=470$ for several methods, with $A=9$ and $\tau_t=5\%$ leading to $\overline{w}_e=19$ and $\tau_e=4.0\%$.
    The Euclidean error is respectively of $9.9$, $10.7$, $10.7$, $18.1$, $6.0$, and $11.5$ for SAX, 1d-SAX, SFA, ABBA, ASTRIDE, and FASTRIDE respectively.
    Note that SFA, ASTRIDE, and FASTRIDE are fitted on the whole training set (and not on the displayed signal only).
    All displayed signals are truncated (so that their length is $\left\lfloor n/w \right\rfloor w$).
    }
    \label{fig:reconstruction_signal_example}
\end{figure}

\paragraph{Comparison of the methods on a single signal}
Figure~\ref{fig:reconstruction_signal_example} gives an example of reconstruction of a single signal from the UCR Time Series Classification Archive, for several methods.
Contrary to ASTRIDE, the change-points from ABBA are not exact but approximated from the cluster centers. Thus, they are not precise in the reconstruction phase, which explains why ABBA behaves better with the DTW error, which allows for time-shifts, than with the Euclidean error.
Note that the authors of ABBA emphasize that their method does not focus on approximating the signal values at the exact timestamps, but rather on capturing the overall behavior.
SFA tends to provide accurate global -- rather than local -- reconstruction: as shown in Figure \ref{fig:reconstruction_signal_example}: depending on the data set, this property can be an advantage or a drawback.
Regarding ASTRIDE, we can see that the segmentation phase allows us to focus on the phenomenon of interest in the signal, thus to devote more memory to the encoding of salient events.

\subsection{Computational complexity}
\label{sec:discussion}

This section describes an important characteristic of the methods: the computational cost.
The processing times of the different methods are compared on the 1-NN classification task applied to the ECG200 data set from the UCR Time Series Classification Archive, and are reported in Table~\ref{tab:processing_time}.
We ran the experiments using Python 3.10.6 on a laptop under macOS 13.0.1 with Apple M1 Chip 8-Core CPU and 7-Core GPU.
All methods mentioned in Section \ref{subsec:bench_classif} are compared (SAX, 1d-SAX, ASTRIDE, and FASTRIDE).
For SAX, both our implementation and \texttt{tslearn} are tested.
Two durations are reported: the time to compute the symbolization for all time series in the data set, and the time to perform the actual 1-NN classification from the symbolized time series.
\begin{table}[hbtp]
    \centering
    \begin{minipage}{0.9\linewidth}
    \caption{Processing times on the symbolization and 1-NN classification on the ECG200 data set (UCR Time Series Classification Archive) composed of 100 training signals and 100 test signals of length $n=96$, with $w=10$ and $A=9$.
    }
    \label{tab:processing_time}
    \centering
    \medbreak
    \begin{tabular}{@{}lp{0.25\linewidth}p{0.25\linewidth}@{}}
    \hline
    Method & Symbolization processing time (s) & 1-NN classification processing time (s) \\
    \hline
    SAX & $0.27$ & $0.08$ \\
    SAX (\texttt{tslearn}) & $0.02$ & $0.11$ \\
    1d-SAX (\texttt{tslearn}) & $0.42$ & $0.21$ \\
    ASTRIDE & $0.30$ & $0.17$ \\
    FASTRIDE & $0.26$ & $0.07$ \\
    \hline
    \end{tabular}
    \end{minipage}
\end{table}

First, as expected, the ASTRIDE symbolization is more time-consuming than the non-adaptive ones (SAX for example).
An important remark is that the temporal segmentation is relatively fast: the computation times for ASTRIDE and its variant FASTRIDE without adaptive segmentation are quite similar.
1d-SAX is more expensive because it takes into account the mean as well as the slope, then has to combine them.

Second, as far as the classification step is concerned, it appears that the computation of D-GED for ASTRIDE is more expensive than for FASTRIDE because of the replication of the symbolic sequences which makes them longer.
For ASTRIDE, using the normalized segment lengths instead of the raw segment lengths helps making the replicated symbolic sequences shorter, but a gap remains in comparison to FASTRIDE.
Several improvements could further lower the computation time, such as using more advanced general edit distances (e.g., Weighted Symbols-Based Edit Distance~\cite{2010_Barat_WeightedED}), which are optimized to deal with redundant series of symbols.

\section{Conclusion}\label{sec:conclusion}

We have introduced a new symbolic representation of time series, ASTRIDE, along with its accelerated variant FASTRIDE, as well as a novel similarity measure on symbolic sequences, D-GED.
ASTRIDE is the only symbolic representation offering adaptive discretization on both the time and amplitude dimension, at the scale of a data set, while having a compatible similarity measure and a reconstruction procedure that is memory-efficient.
Hence, ASTRIDE combined with D-GED alleviates the main drawbacks of existing symbolic representations.
ASTRIDE uses change-point detection of mean-shifts instead of uniform segmentation, and adaptive quantization using quantiles in place of fixed Gaussian bins.
Moreover, D-GED is based on the general edit distance which relies on the quantiles and deals with substitution, deletions and insertions, which are not handled by the MINDIST similarity measure.
In addition, thanks to the multivariate change-points and the quantiles learned from all signals in the data set, the dictionary of symbols is shared across all signals, thus reducing memory usage.

Our experiments show the quality of our symbolic representations.
Indeed, both ASTRIDE and FASTRIDE give better accuracies than SAX and 1d-SAX on the classification task, for a same word length.
This performance is mainly due to the adaptive quantization based on quantiles and the D-GED similarity measure.
For the reconstruction task, FASTRIDE and especially ASTRIDE give better errors than SAX, 1d-SAX and ABBA, for a same memory usage ratio.
On the reconstruction task, the adaptive segmentation is particularly relevant, thus using ASTRIDE rather than FASTRIDE seems more appropriate.

Further works will investigate the possibility of using ASTRIDE for classifiers that are built on top of symbolic representations, for example using ASTRIDE instead of SFA in the symbolization step of BOSS.
We also plan to extend our approach to multivariate time series.
We intend to further improve the computation time of D-GED by using more advanced string distances that are specifically designed to deal with redundant series of symbols.
Finally, this new representation could also be employed for more advanced data mining tasks, such as pattern mining or anomaly detection.

\subsection*{Acknowledgments}

This work is supported by a public grant overseen by the French National Research Agency (ANR) through the program UDOPIA, project funded by the ANR-20-THIA-0013-01.


\end{document}